\documentclass[10pt,twocolumn,letterpaper]{article}

\usepackage{cvpr}
\usepackage{times}
\usepackage{epsfig}
\usepackage{graphicx}
\usepackage{amsmath}
\usepackage{amssymb}
\usepackage{lipsum}
\usepackage{multirow}
\usepackage{subfig}

\usepackage[breaklinks=true,bookmarks=false]{hyperref}

\cvprfinalcopy 

\def\cvprPaperID{****} 

\begin{document}

\title{Feeding Hand-Crafted Features for Enhancing the
Performance of Convolutional Neural Networks}

\author{Sepidehsadat Hosseini\\
Seoul National University\\
sepid@ispl.snu.ac.kr
\and
Seok Hee Lee\\
Seoul Nat'l Univ.\\
seokheel@snu.ac.kr
\and
 Nam Ik Cho \\
Seoul National University\\
nicho@snu.ac.kr
}

\maketitle

\begin{abstract}
Since the convolutional neural network (CNN) is believed to find right features for a given problem, the study of hand-crafted features is somewhat neglected these days. In this paper, we show that finding an appropriate feature for the given problem may be still important as they can enhance the performance of CNN-based algorithms. Specifically, we show that feeding an appropriate feature to the CNN enhances its performance in some face related works such as age/gender estimation, face detection and emotion recognition. We use Gabor filter bank responses for these tasks, feeding them to the CNN along with the input image. The stack of image and Gabor responses can be fed to the CNN as a tensor input, or as a fused image which is a weighted sum of image and Gabor responses. The Gabor filter parameters can also be tuned depending on the given problem, for increasing the performance. From the extensive experiments, it is shown that the proposed methods provide better performance than the conventional CNN-based methods that use only the input images. 
\end{abstract}

\section{Introduction}

The CNNs are gaining more and more attention as they are successfully applied to many image processing and computer vision tasks, providing better performance than the non-CNN approaches. Face related tasks are not the exceptions, for example, the CNNs in~\cite{facedetect_cascade,faceness,MTCNN} provide better face detection performance than the conventional methods such as Haar-like feature based face detector \cite{viola}, local binary pattern (LBP) based method \cite{LBP} and deformable part model based ones \cite{dpm_2,dpm_1}. In the case of age/gender classification, the CNN estimators \cite{levi, hybridmulti} give more accurate results than the method based on the bio-inspired features (BIF) \cite{bif}, which is one of the best methods among the non-CNN approaches. 

Most of CNNs from low to high-level vision problems use the image (not the features) as the input, and they  learn and extract the features from the training data without human intervention. In this paper, we show that feeding some effective hand-crafted features to the CNN, along with the input images, can enhance the performance of CNN at least in the case of some face related tasks that we focus on. In other words, enforcing the CNN to use the domain knowledge can increase the performance or can save the computations by reducing the depth. To be specific with the age/gender estimation problem, since the most important features are the angle and depth of the wrinkles in our faces, we believe that the bio-inspired multi-scale Gabor filter responses \cite{bif} are the right features for this problem. Hence, we propose a method to get the benefits of BIF, together with the features that are learned by the CNN with the input images. Precisely, we extract several Gabor filter responses and concatenate them with the input image, which forms a tensor input like a multi-channel image. The tensor input can be directly fed to the CNN, like we feed the multi-channel image to the CNN. In addition to this scheme, we let the first layer of the CNN to be a $1 \times 1$ convolution such that a matrix is obtained at the first layer, which is actually a weighted sum of the input image and Gabor responses. This can also be considered a fusion of input image and filter bank responses, which looks like an image with enhanced trextures, and the fused image is fed to the CNN. 

Analysis of feature maps from some of convolution layers shows that the wrinkle features and face shapes are more enhanced in our CNN than the conventional one that uses only the pixel values as the input. As a result, the accuracy of age/gender estimation is much improved compared to the state-of-the-art image-domain CNNs \cite{levi,hybridmulti}. Moreover, we test our approach on face detection and emotion recognition and also obtain some gains over the existing CNN based methods \cite{facedetect_cascade,faceness,MTCNN}. In other tasks where some of the hand-crafted features are apparently effective, we hope that feeding such features along with the image may bring better results.

\section{Related work}

\begin{figure}[h!]
  \includegraphics[width=\linewidth]{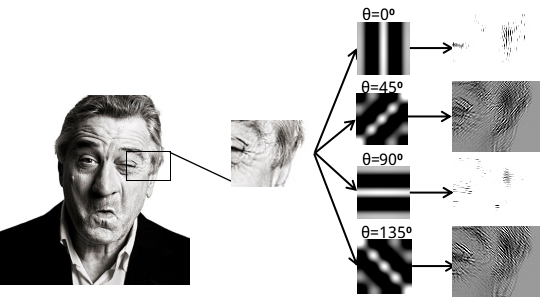}
  \caption{Demonstration of Gabor filter bank responses with kernel size = 5 applied to an image. 
  Responses for four orientations ($\theta = 0,\pi/4,\pi/2,3\pi/4$) are shown.}
  \label{fig:gabor}
\end{figure}

\noindent {\bf Gaobr filters.} Nobel prize winners Hubel and Wiesel discovered that there are simple cells in the primary visual cortex, where its receptive field is divided into subregions that are layers covering the whole field \cite{gabor_bio}. Also in \cite{gabor_math}, Petkov proposed the Gabor filter, as a suitable approximation of  mammal's visual cortex receptive field. The 2D Gabor filter is a Gaussian kernel function adjusted by a sinusoidal wave, consisting of both imaginary and real parts where the real part can be described as:
 \begin{equation}
g_{\lambda,\theta,\sigma,\gamma}(x,y)=\exp\left(-\frac{x'+\gamma y'^2}{2\sigma^2}\right)\cos \left(2\pi \frac{x'}{\lambda}+\phi \right)  \label{eq:gabor}
\end{equation}
where $x'=x \cos\theta + y \sin\theta$, 
$y'=-x \sin\theta + y \cos\theta$,
and  $\lambda$, $\theta$, $\phi$ , $\gamma$ and $\sigma$ are the wavelength of the real part of Gabor filter kernel, the orientation of the normal to the stripes of function, phase offset, spatial ratio and standard
deviation of the Gaussian envelope  representatives respectively.
Fig.~\ref{fig:gabor} is an example of Gabor filter response to a face image, which shows that they find the textures that correspond to the given $\theta$ very well. Hence the Gabor filter responses have been used in the applications where the (orientational) textures play an important role such as fingerprint recognition \cite{gabor_fingerprint}, face detection \cite{gabor_facedetect}, facial expression recognition \cite{gabor_emotion}, age/gender estimation \cite{bif}, text segmentation \cite{gabor_txt}, super resolution \cite{gabor_super}, and texture description. 

\vspace{0.3cm}
\noindent {\bf Age/Gender Estimation.} Predicting the age of a person from a single image is one of the hardest tasks, which even humans sometimes have difficulties in doing that. The reason is that aging depends on several factors such as living habits, races, genetics, etc. The studies without using the CNN are well summarized in Fu et al.'s survey \cite{survey_age}. Recent works are mostly based on the CNN, for some examples, Levi and Hassner's work \cite{levi} was the first to adopt the CNN for age/gender estimation, and Xing et al. \cite{hybridmulti} considered the influence of race and gender by proposing a multi-task network.

\vspace{0.3cm}
\noindent {\bf Face detection.} There are a large number of face detection methods, as it is also a very important topic. For details, refer to a complete survey on face detection done by Zafeiriou et al.\cite{survey_facedetect}. Like other computer vision problems, the CNNs are now effectively used for face detection \cite{ faceness,widerface,multiscale}. 

\vspace{0.3cm}
\noindent {\bf Facial Expression Recognition}
Emotion classification is a relatively young and complicated task among many face-related tasks. Since the facial expression recognition (FER) plays an important role in human-machine interaction, recently more researches are being performed on this subject. For some examples of conventional methods, Tang used support vector machine (SVM) for this problem \cite{Tang}. Ionwscu et al. also used SVM to improve Bag of Visual words (BOW) approach \cite{bow}. Hassani et al. used the advantage of facial landmarks along with CNNs \cite{hassani}. More recent studies are focused on using the CNNs for the FER \cite{yu,tal, statecnn}.


\section{Preparation of Input}

We attempt our approach to several face related works such as age/gender estimation, face detection, and emotion recognition. Each of them needs different CNN architecture, but they are all fed with the Gabor filter responses as the input along with the image. As can be seen from the eq.~(\ref{eq:gabor}), there are several parameters which induce different filter responses. In all the applications, we prepare eight filter banks by combining the cases of four $\theta ={0,\pi/4 , \pi/2 ,3\pi/4} $ and two $\phi= {0, \pi/2} $. The rest of parameters $\sigma$, $\lambda$ and $\gamma$ are changed depending on the application. For the age and gender estimation problem, we set $\sigma = 2$, $\lambda = 2.5$, and $\gamma=0.3$. 

\begin{figure*}[t!]
\centering

     \includegraphics[ keepaspectratio, width=\textwidth]{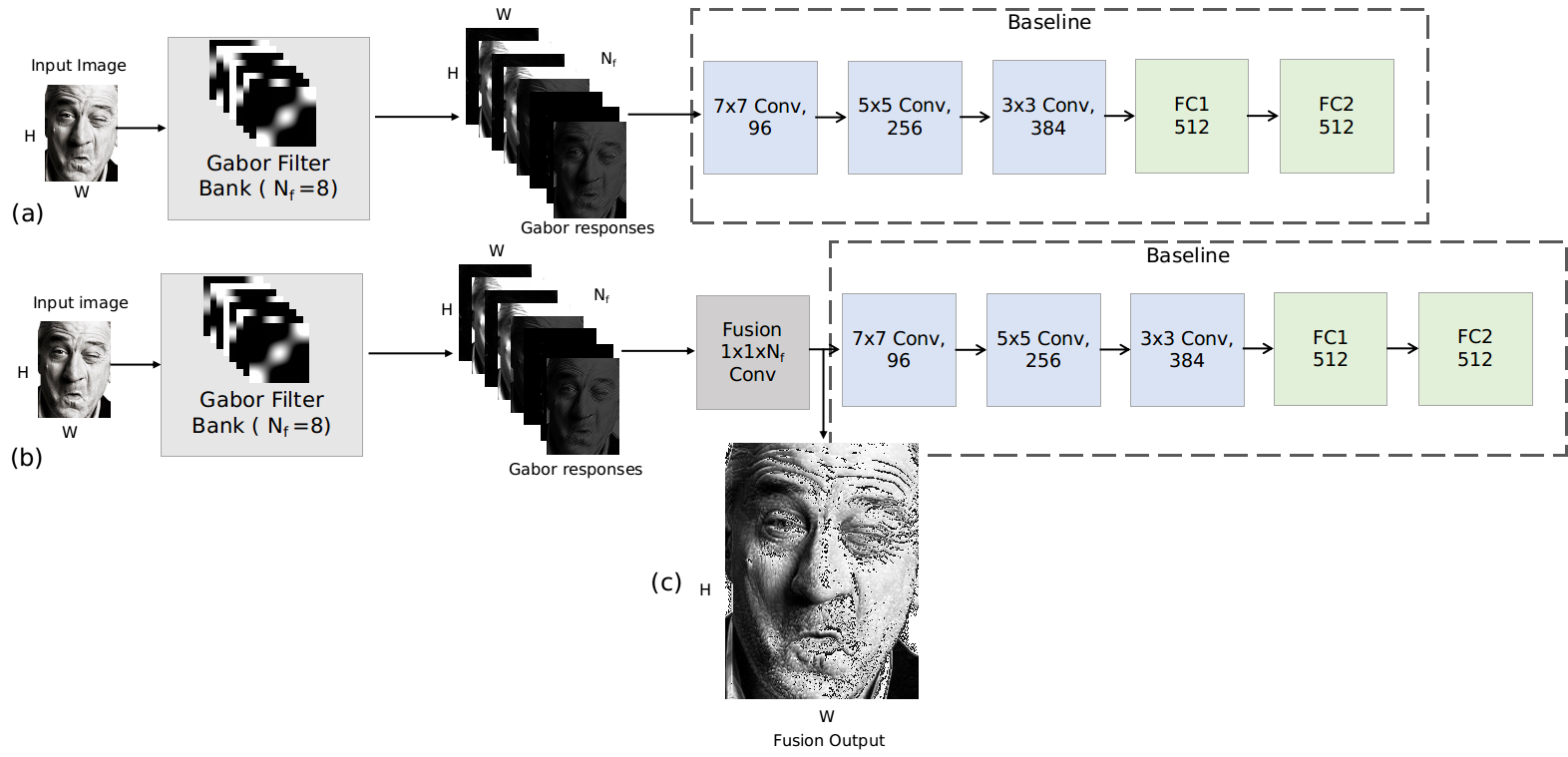}

  \caption{Illustration of two input feeding methods. (a) The tensor input is directly fed to the CNN, (b) The tensor input is fused to be an image and fed to the CNN. (c) Example of a fused image which is the weighted sum of image and Gabor responses.}
 
  \label{fig:input}
\end{figure*}

Let $N_f$ (=8 in all the experiments in this paper as stated above) be the number of Gabor filters, and let $F_g^k$ be the response of $k$-th Gabor filter. Normally, we may just concatenate the input image and $N_f$ responses as $W \times H \times (N_f+1)$ tensor input to a CNN as illustrated in Fig.~\ref{fig:input}(a). On the other hand, we may consider fusing the input and Gabor responses as a single input and feed the matrix to the CNN as shown in Fig.~\ref{fig:input}(b). The figure also shows that fusing the input image and Gabor responses can be interpreted as convolving the $W \times H \times (N_f+1)$ tensor input with $1 \times 1 \times (N_f+1)$ filter. If we denote the coefficients of this filter as $[w_i, \; w_1, \; w_2, \cdots, w_{N_f}]$, where $w_i$ is multiplied to the input image and the rest are multiplied to Gabor responses, then the fused input is represented as
\begin{equation}
F^{in} =w_i I + \sum_{k={1}}^{N_{f}} w_{k} F_g^k 
\end{equation}
which is similar to the weighted fusion method in \cite{fusion},\cite{hybridmulti}.
Fig.~\ref{fig:input}(c) is an example of fused input, which can be considered a ``wrinkle-enhanced'' image.
Both of concatenation and fusion approaches inject the Gabor responses as the input to the CNN. From the extensive experiments, the fusion approach in  Fig.~\ref{fig:input}(b) shows slightly better performance (about 1\%p increase in the case of gender estimation and similarly to other tasks) while requiring slightly less number of parameters. 

\section{Networks for face related problems}

We apply the Gabor responses to the CNNs for the age/gender estimation, face detection, and emotion recognition problems in the following subsections. At each subsection, we show that the performance is improved by feeding the Gabor responses as compared to the case of feeding only the image input.


\subsection{Age/Gender classification}
\renewcommand{\thesubsubsection}{\Alph{subsubsection}}

\subsubsection{Network} 
\quad The gender estimation is just a binary classification, while the age estimation is implemented as a classification or regression problem. In the case of age estimation as a classification problem (segmenting the age into several ranges), the network shown in Fig.~\ref{fig:input}(a) or (b) is used. Each convolution block consists of convolution layer, Relu, and Max pooling, and each fully connected block consists of fully connected layer, Relu and drop-out with the drop ratio 0.5.

\subsubsection{Dataset description} 
\quad We perform age classification on two popular datasets, Adience \cite{adiencedataset} and Gallagher dataset \cite{gallagher_dataset}. Both are from flickr.com, including the pictures with large variations in poses, appearances, lighting condition, unusual facial expressions, etc. Adience has approximatively 26K images of 2k subjects in 8 classes (0-2, 4-6, 8-13, 15-20, 25-32, 38-42, 48-53, 60+), Gallagher  dataset has 5K images with 28K labeled faces, being divided into 7 classes (0-2, 3-7,8-12, 13-19, 20-36, 37-65, 66+).  
For gender estimation, we used Adience and CASIA\_Webface \cite{casia}. It has 450K images of 10K subjects, which is obtained from the pictures on IMDB and most of the pictures in the dataset are celebrities.

\subsubsection{Test and result} 
\quad We perform the experiments based on the standard five-fold, subject-exclusive cross-validation protocol for fair comparison. Table \ref{table:1} shows the results for age estimation, where GT\_CNN means our method that use Gabor responses as tensor input and GF\_CNN as fused input. It can be observed that GF\_CNN is slightly better than GT\_CNN as stated previously, and the GF\_CNN outperforms the existing methods by at least 3.1 $\%$p on Aidence dataset and 1.3$\%$p on Gallagher dataset. 

For gender estimation, our method outperforms all the other ones on Adience as shown in Table \ref{table:2}.  The Table also shows that the proposed network shows almost the same performance as VGG hybrid on Webface dataset, while it has ten times less number of parameters than the VGG.

\begin{figure}[htbp]

\begin{center}
(a)\quad{\includegraphics[height=1cm,width=.4\textwidth ]{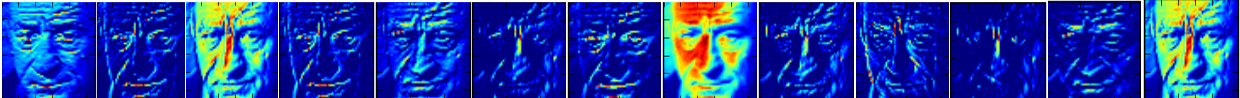}}
\\(b)\quad{\includegraphics[height=1cm,width=.4\textwidth]{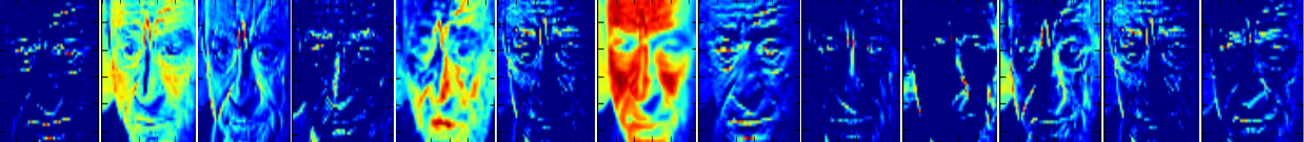}} 

\end{center}
\caption{Comparison of feature maps after the first convolution layer in two networks. Features from (a) GF\_CNN and (b) original CNN with image input.}

  \label{fig:feature}
\end{figure}

For the analysis of the effects of feeding the Gabor responses, we compare some feature maps in Fig.~\ref{fig:feature}. 
Specifically, Fig.~\ref{fig:feature}(a) shows the feature maps from our GF\_CNN and  Fig.~\ref{fig:feature}(b) from the CNN with only image input at the same layer. It can be seen that the features from the GF\_CNN contain more strong facial features and wrinkle textures than the original network, which is believed to be the cause of better performance.

\begin{table}[htbp]
\caption{Age estimation (classification) results on Adience \& Gallagher datasets.} 
\begin{center}
\begin{tabular}{|l | {c} |r|}
\hline
 Method &    Adience &  Gallagher \\\hline  LBP \cite{age_1} & 41.1 & 58.0 \\ LBP+FPLBP+Droupout 0.8 \cite{age_1} & 45.1 & 66.6 \\ Eidinger \cite{adiencedataset} & 45.1 & N.A.  \\
 Best from Levi \cite{levi} & 50.7 & N.A. \\
 Resnet\cite{resnet} & 52.2 & 68.1 \\
  PTP \cite{dAPP} & 53.27 & 68.6 \\
 DAPP \cite{dAPP} & 54.9 & 69.91 \\
 GT\_CNN[Ours]  & { 57.2} & { 69.1}\\
 GF\_CNN[Ours]  & {\bf 59.3} & {\bf 71.4}\\
\hline
\end{tabular}
\end{center}

\label{table:1}
\end{table}
\begin{table}[h]
\caption{Gender estimation results on Adience \& Webface datasets.}
\begin{center}
\begin{tabular}{|l | {c} |r|}
\hline
 Method &    Adience &  Webface \\
 \hline 
  BIF \cite{bif} & N.A & 79.3 \\ 
 Eidinger \cite{adiencedataset} & 77.8 & N.A.  \\
 Best from Levi \cite{levi} &  86.8 & N.A. \\
 Resnet\cite{resnet} & 88.5 & 89.2  \\
$Net^{VGG}_{Hybrid} $\cite{hybridmulti} & N.A & {\bf 92.3} \\
 GT\_CNN[Ours]  & { 89.2 } & { 91.0}\\
 GF\_CNN[Ours]  & {\bf 90.1 } & { 92.1}\\
\hline
\end{tabular}
\end{center}

 \label{table:2}

\end{table}

\subsection{Age regression }


\subsubsection{Network} 
\begin{figure}[h!]
 \includegraphics[height=1.6cm ,width=\linewidth]{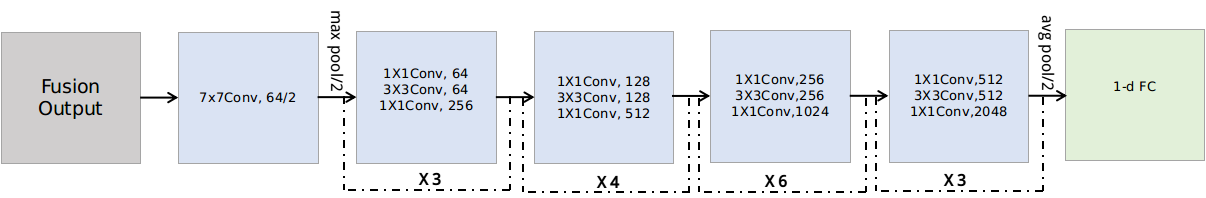}
  \caption{Age regression network architecture.}
  \label{fig:regression}
\end{figure}

\quad 
Age estimation can also be implemented as a regression problem when we wish to tell a person's exact age, rather than as a classification problem which tells the range (class) of ages. We use the network shown in Fig.~\ref{fig:regression} for this problem. One of the main differences between the age classification and regression problem is that they need different loss functions. 
For the classification problem above, we use the Softmax loss defined as: 
\begin{equation}
L(x)=-\frac{1}{N}\sum_{i=1}^{N}Y_{iy_i} \log p_{iy_i}
\end{equation} 
where $N$ is the number of classes,  $Y_{iy_i}$ is the one-hot encoding of sample's age label, and $p_{iy_i}$ is the $y_{i}$-th element of predicted probability vector for $x_{i}$.
For the regression, we use Mean Squared Error (MSE) or Mean Absolute Error (MAE) as the loss function. To be precise, the MAE is defined as
\begin{equation}
L(x)=-\frac{1}{M}\sum_{i=1}^{M}|\hat{y_{i}} - y_i|
\end{equation}
where $M$ is the maximum age that we set, and $\hat{y_i}$ is the estimate of true age $y_i$.

\subsubsection{Dataset description} 
\quad
For age regression task, we perform the experiments on two widely used datasets for age estimation in literature. We choose CASIA-Webface dataset ~\cite{casia} as it consists of a large amount of pictures, and also we used FG-Net database which contains 1002 images of 82 subjects, where subjects' ages range from 0 to 69.

\subsubsection{Test and result} 
\quad 
We used four-fold cross-validation protocol for Webface dataset and the Leave-One-Person-Out (LOPO) test strategy  while working on FG-Net because the number of pictures in FG-Net is small. Table~\ref{table:3} shows the result of age estimation. It can be seen that our network shows better performance than the state of the art method.

\begin{table}[htbp]
\caption{Age estimation error on Adience and Gallagher datasets. GF\_CNN$_{resent}$ means that we use residual learning.} 
\begin{center}
\begin{tabular}{|l | {c} |r|}
\hline
 Method &    Casia Webface Dataset & FG-net \\\hline 
 BIF\cite{bif}  & 10.65 &4.77 \\ 
   RF\cite{RF}  & 9.38 & 4.21 \\
   EBIF\cite{ebif}  & N.A. & 3.17  \\ $Net^{VGG}_{Hybrid}$ & 5.75 & N.A. \\GF\_ CNN[ours] & 5.83 & 3.13  \\  GT\_CNN$_{resent}$[ours] &   5.66 & 3.15\\ GF\_CNN$_{resent}$[ours] &  {\bf 5.61} & {\bf 3.08} \\
\hline
\end{tabular}
\end{center}
\label{table:3}
\end{table}

\subsection{Face detection}

\subsubsection{Network} 
\begin{figure}[htbp]
\centering
 \includegraphics[ height=5.6cm, width=0.5\textwidth]{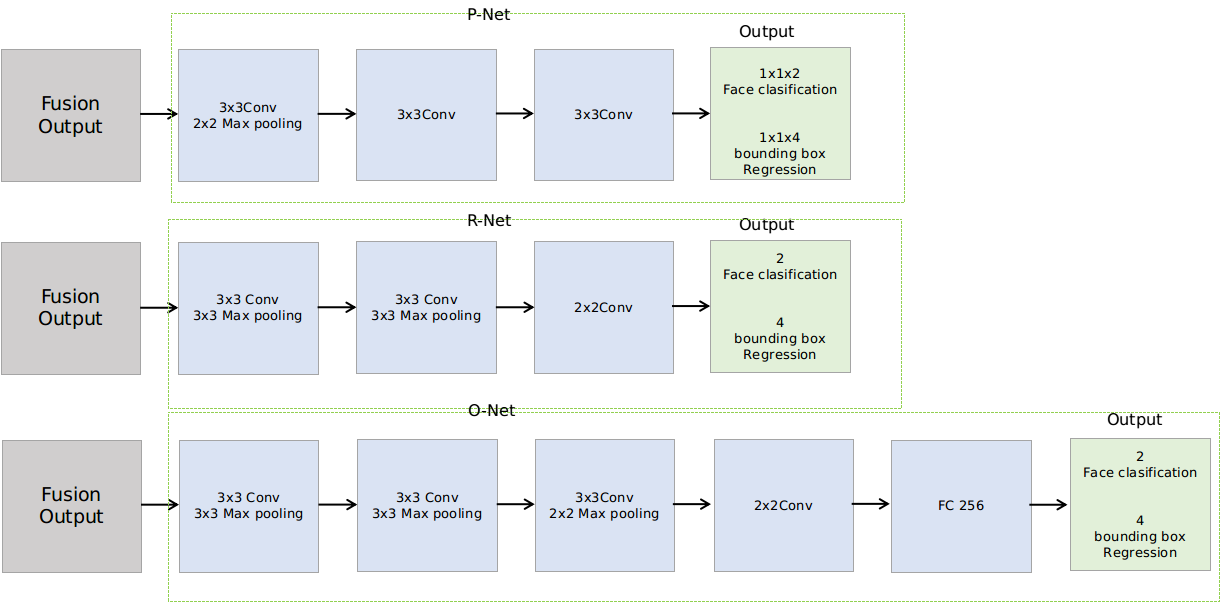}

  \caption{Illustration of three stages of face detection network architecture (GP-Net, GR-Net ,GO-Net).}
 
  \label{fig:detection}
\end{figure}

\quad Our face detector is a three-stage cascaded CNN which is the same as Zhang et al.'s network \cite{MTCNN}, except that we use the fusion of input and Gabor responses as shown in Fig.~\ref{fig:detection}.
At stage 1, which is called P-Net, possible facial windows along with their bonding box regression vectors are obtained. Then the bounding boxes are calibrated, and the highly overlapped ones are merged to others using non-maximum suppression (NMS). In the second and third stages (called R-Net and O-Net respectively) the candidates are refined again using the calibration and NMS. For all these three step networks we feed our Gabor fusion image. 

About the Gabor filter parameters,  it is noted that finding the facial components such as nose, mouse, eyes, etc. are more important than the relatively straight and sometimes long wrinkles that were important in the previous age/gender estimation. Hence we reduce the kernel size of Gabor filter and also the parameters $\sigma$, $\lambda$ and $\gamma$ to 0.75, 2, and 0.05 respectively.  

\subsubsection{Dataset description} 
\quad In this section, we evaluate our network on Face Detection Dataset and Benchmark (FDDB) \cite{fddb} which contains 2,845 images with 5K annotated faces taken in the wild. There are two types of evaluation available on FDDB: {\it discontinuous score} which counts the number of detected faces versus the number of false positives, and {\it continuous score} which evaluates how much is the overlap of bounding boxes on the faces between the ground truth and detected.

\subsubsection{Test and result} 

\begin{figure*}[htbp]

\begin{center}
{\includegraphics[height=4.5cm ,width=.4\textwidth ]{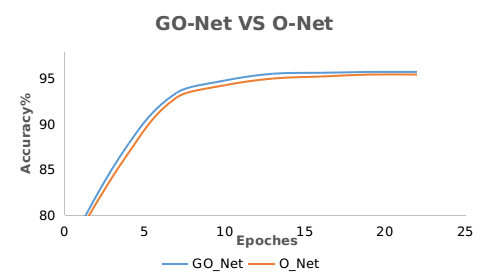}}
{\includegraphics[height=4.5cm,width=.4\textwidth]{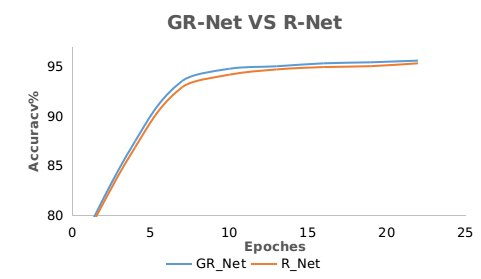}} \\(a) \qquad \qquad \qquad \qquad \qquad \qquad \qquad  \quad \quad \qquad \qquad (b)\\

{\includegraphics[height=4.5cm,width=.4\textwidth]{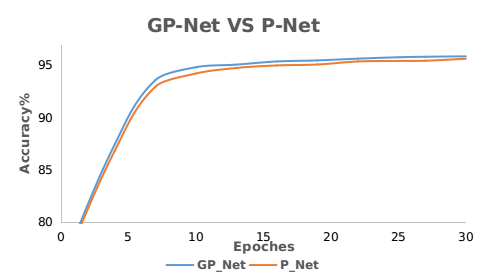}}
{\includegraphics[height=4.5cm,width=.4\textwidth]{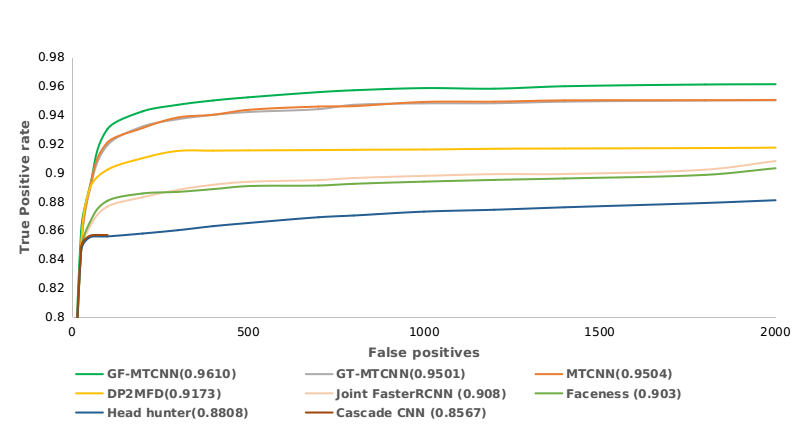}}
 \\(c) \qquad \qquad \qquad \qquad \qquad \qquad \qquad  \quad \quad \qquad \qquad (d)\\
\end{center}
\caption{(a)-(c) Comparision between the three stages of MTCNN \cite{MTCNN} (in orange) and our method (in Green). (d) Comparison of our performance with MTCNN \cite{MTCNN}, DP2MDF\cite{dpm_1}, cascade CNN \cite{facedetect_cascade}, Faceness\cite{faceness}, Joint fasterRCNN \cite{jointcascade_fast} and head hunter \cite{headhunter}, where the numbers in the parentheses are the area under curve.}
\label{gmtcnn}
\end{figure*}

\quad For the bounding box regression and face classification, we use the same loss as \cite{MTCNN}. 
Specifically, we use cross-entropy loss:
\begin{equation}
L_{i}^{det}(x_{i}) =-( y_{i}^{det}log(p_{i})+(1-y_{i}^{det})(1-log(p_{i})))  
\end{equation}
where $p_{i}$ is the probability of $x_{i}$ being a face and $y_{i}^{det}$ is the ground truth. 
For the bonding box we use: 
\begin{equation}
 L_{i}^{B.Box}(x_{i})=||\hat{y}_{i}^{B.box}-y_{i}^{B.box}||_2^{2} 
\end{equation}
where $\hat{y}_{i}^{B.box}$ and$ y_{i}^{B.box}$ are the network output and ground truth respectively. 
Table~\ref{table:4} shows that we can get better performance with almost same number parameters as MTCNN. Figs.~\ref{gmtcnn}(a)-(c) show in all three stages using hand crafted features can improve the performance and help increase the network convergence speed. To evaluate our face detection method we compare our method with other six sate- of-the-art methods on FDDB and our method outperform all of them as shown in Fig.~\ref{gmtcnn}(d). At last, we compare our method's run time with other CNN based methods and results are in ~\ref{table:5} as it can be seen while purposed method has better performance than MTCNN and cascade CNN it is almost as fast as them.

\begin{table}[htbp]
\caption{Comparison of Validation Accuracy of Ours, CascadeCNN and MTCNN. } 
\begin{center}

\begin{tabular}{|l | {c} |r|}
\hline
 Group &    CNN &  Validation Accuracy \\\hline  
  & 12-Net\cite{facedetect_cascade} & 94.4$\%$ \\  
Group1  & P-Net\cite{MTCNN}& 94.6$\%$ \\
 & GP-Net[ours]. & 94.83$\%$\\\hline 
  
  &  24-Net\cite{facedetect_cascade} & 95.1$\%$ \\  
Group2  &  R-Net\cite{MTCNN} & 95.4$\%$ \\
  & GR-Net[ours] & 95.61$\%$\\\hline 
  & 48-Net\cite{facedetect_cascade} &93.2$\%$ \\  
Group3  &  O-Net\cite{MTCNN} & 95.4$\%$ \\
 & GO-Net[ours] & 95.72$\%$\\\hline   

\end{tabular}
\end{center}
\label{table:4}
\end{table}

\begin{table}[htbp]
\caption{Runtime Comparison on the same GPU.} 
\begin{center}

\begin{tabular}{|l | r|}
\hline
 Method &      Speed \\\hline  
 Faceness \cite{faceness} &20 FPS \\
MTCNN \cite{MTCNN}  & 99 FPS \\
Cascade CNN \cite{facedetect_cascade}  & 100 FPS \\
GF-MTCNN[Ours] & 99 FPS\\
\hline

\end{tabular}
\end{center}

\label{table:5}
\end{table}

\subsection{Facial expression recognition}

\renewcommand{\thesubsubsection}{\Alph{subsubsection}}

\subsubsection{Network} 

\begin{figure}[htbp]
\centering
 \includegraphics[ height=1.3cm , width=0.5\textwidth]{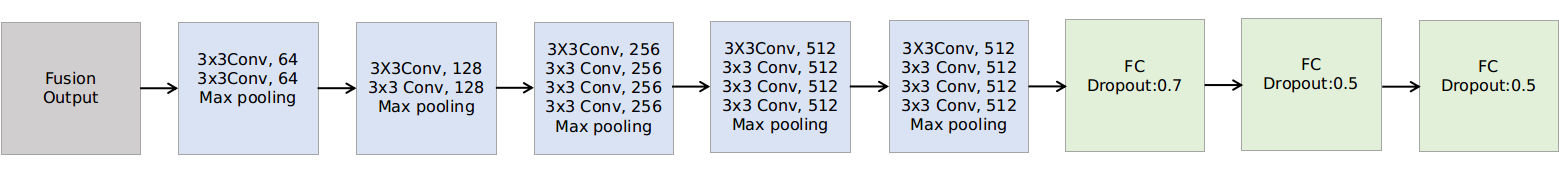}

  \caption{Illustration GF-VGG network for the FER.}
 
  \label{fig:vgg}
\end{figure}

The baseline network for FER is VGG-19 \cite{vgg}, and we just add one more drop out after the last fully connected layer to decrease the overlapping, as shown in Fig.~\ref{fig:vgg}. For the FER, we think that the wrinkles again play an important role here, and hence that we set the bandwith larger than the previous case, specifically set $\sigma = 1.4$. Also, $\lambda$ becomes large to 2.5, and set $\gamma =0.1$.

\subsubsection{Dataset description} 
We evaluate our network on FER 2013 dataset \cite{fer}, which is being labeled  in seven classes(0=Angry, 1=Disgust, 2=Fear, 3=Happy, 4=Sad, 5=Surprise, and 6=Neutral). It contains about 32K images, 28.5K for training and 3.5K for the test. 

\subsubsection{Test and result} 
Table~\ref{table:6} shows our result, where we compare our results with the FER 2013 competition winners and other state of the art methods. It can be seen that our network shows better performance than others. While VGGNet can also reach to 69.8 $\%$, adding our fusion module at the input of the network can increase the performance by 2.098$\%$p. 

\begin{table}[htbp]
\caption{Results of FER.} 
\begin{center}
\begin{tabular}{|l | {c} |r|}
\hline
 Method &   Accuracy on FER 2013\\\hline  
Radu + Marius + Cristi \cite{bow}  & 67.484$\%$ \\
Unsupervised\cite{fer} &69.267$\%$\\
Maxim Milakov \cite{fer} & 68.821$\%$ \\
SVM \cite{Tang} & 71.162$\%$ \\
VGGNet \cite{vgg} & 69.08$\%$ \\
GF-VGGNet[Ours] & {\bf 72.198$\%$} \\
\hline   
\end{tabular}
\end{center}
\label{table:6}
\end{table}

\section{Conclusion}
Most of CNNs for image understanding use the image as the input, with the belief that the CNN will automatically find the appropriate features from the data. In this paper, we have shown that feeding appropriate hand-crafted features can lead to the improved results. Hence the domain knowledge and study of appropriate features are important for improving the CNN-based algorithms. Specifically, we have shown that feeding the Gabor filter response to the CNN leads to better performances in face related problems such as age/gender estimation, face detection, and emotion recognition. We hope there can be more applications that can be benefited by our approach, i.e., there can be more CNN-based image processing and vision algorithms that can have gains by taking the appropriate features as the input.


\bibliographystyle{ieee}
\bibliography{egbib}

\begin{thebibliography}{10}\itemsep=-1pt

\bibitem{LBP}
O.~Bilaniuk, E.~Fazl-Ersi, R.~Laganière, C.~Xu, D.~Laroche, and C.~Moulder.
\newblock Fast lbp face detection on low-power simd architectures.
\newblock In {\em Proceedings of the IEEE Conference on Computer Vision and
  Pattern Recognition Workshops}, pages 616--622, 2014.

\bibitem{ebif}
M.~E. Deeb and M.~El-Saban.
\newblock Human age estimation using enhanced bio-inspired features (ebif).
\newblock In {\em ICIP}, 2010.

\bibitem{adiencedataset}
E.~Eidinger, R.~Enbar, and T.~Hassner.
\newblock Age and gender estimation of unfiltered faces.
\newblock {\em IEEE Transactions on Information Forensics and Security},
  9(12):2170--2179, 2014.

\bibitem{gallagher_dataset}
A.~C. Gallagher and T.~Chen.
\newblock Understanding images of groups of people.
\newblock In {\em Computer Vision and Pattern Recognition, 2009. CVPR 2009.
  IEEE Conference on}, pages 256--263. IEEE, 2009.

\bibitem{fer}
I.~Goodfellow, D.~Erhan, P.-L. Carrier, A.~Courville, M.~Mirza, B.~Hamner,
  W.~Cukierski, Y.~Tang, D.~Thaler, D.-H. Lee, Y.~Zhou, C.~Ramaiah, F.~Feng,
  R.~Li, X.~Wang, D.~Athanasakis, J.~Shawe-Taylor, M.~Milakov, J.~Park,
  R.~Ionescu, M.~Popescu, C.~Grozea, J.~Bergstra, J.~Xie, L.~Romaszko, B.~Xu,
  Z.~Chuang, and Y.~Bengio.
\newblock Challenges in representation learning: A report on three machine
  learning contests, 2013.

\bibitem{gabor_fingerprint}
C.~Gottschlich.
\newblock Curved-region-based ridge frequency estimation and curved gabor
  filters for fingerprint image enhancement.
\newblock {\em IEEE Transactions on Image Processing}, 21(4):2220--2227, 2011.

\bibitem{bif}
G.~Guo, G.~Mu, Y.~Fu, and T.~S.~Huang.
\newblock Human age estimation using bio-inspired features.
\newblock In {\em Computer Vision and Pattern Recognition, 2009. CVPR 2009.
  IEEE Conference on}. IEEE, 2009.

\bibitem{hassani}
B.~Hassani and M.~H. Mahoor.
\newblock Facial expression recognition using enhanced deep 3d convolutional
  neural networks.
\newblock {\em CoRR}, abs/1705.07871, 2017.

\bibitem{resnet}
K.~He, X.~Zhang, S.~Ren, and J.~Sun.
\newblock Deep residual learning for image recognition.
\newblock {\em arXiv preprint arXiv:1512.03385}, 2015.

\bibitem{gabor_facedetect}
L.-L. Huang, A.~Shimizu, and H.~Kobatake.
\newblock Robust face detection using gabor filter features.
\newblock {\em Pattern Recognition Letters}, 26(11):1641--1649, 2005.

\bibitem{gabor_bio}
D.~H. Hubel and T.~Wiese.
\newblock Receptive fields, binocular interaction and functional architecture
  in the cat's visual cortex.
\newblock {\em Journal of Physiology}, 160(1):106,154, 1962.

\bibitem{bow}
R.~T. Ionescu and C.~Grozea.
\newblock Local learning to improve bag of visual words model for facial
  expression recognition.
\newblock 2013.

\bibitem{dAPP}
M.~T.~B. Iqbal, M.~Shoyaib, B.~Ryu, M.~Abdullah-Al-Wadud, and O.~Chae.
\newblock Directional age-primitive pattern (dapp) for human age group
  recognition and age estimation.
\newblock {\em IEEE Transactions on Information Forensics and Security},
  12:2505--2517, 2017.

\bibitem{gabor_emotion}
M.~J.~Lyons, S.~Akamatsu, M.~G. Kamachi, and J.~Gyoba.
\newblock Coding facial expressions with gabor wavelets.
\newblock In {\em Automatic Face and Gesture Recognition, 1998. Proceedings.
  Third IEEE International Conference on}. IEEE, 1998.

\bibitem{fddb}
V.~Jain and E.~Learned-Miller.
\newblock Fddb: A benchmark for face detection in unconstrained settings.
\newblock Technical Report UM-CS-2010-009, University of Massachusetts,
  Amherst, 2010.

\bibitem{levi}
G.~Levi and T.~Hassner.
\newblock Age and gender classification using convolutional neural network.
\newblock In {\em Computer Vision and Pattern Recognition Workshops (CVPRW),
  2015 IEEE Conference on}, pages 34--42. IEEE, 2015.

\bibitem{tal}
G.~Levi and T.~Hassner.
\newblock Emotion recognition in the wild via convolutional neural networks and
  mapped binary patterns.
\newblock In {\em Proc. ACM International Conference on Multimodal Interaction
  (ICMI)}, November 2015.

\bibitem{facedetect_cascade}
H.~Li, Z.~Lin, X.~Shen, J.~Brandt, and G.~Hua.
\newblock A convolutional neural network cascade for face detection.
\newblock In {\em Computer Vision and Pattern Recognition (CVPR), 2015 IEEE
  Conference on}, pages 5325--5334, 2015.

\bibitem{RF}
S.~Li, S.~Shan, and X.~Chen.
\newblock {\em Relative Forest for Attribute Prediction}, pages 316--327.
\newblock Springer Berlin Heidelberg, Berlin, Heidelberg, 2013.

\bibitem{headhunter}
M.~Mathias, R.~Benenson, M.~Pedersoli, and L.~{Van Gool}.
\newblock Face detection without bells and whistles.
\newblock In {\em ECCV}, 2014.

\bibitem{gabor_math}
N.~Petkov.
\newblock Biologically motivated computationally intensive approaches to image
  pattern recognition.
\newblock {\em Future Generation Computer Systems}, 11(4--5):451,465, 1995.

\bibitem{statecnn}
C.~Pramerdorfer and M.~Kampel.
\newblock Facial expression recognition using convolutional neural networks:
  State of the art.
\newblock {\em CoRR}, abs/1612.02903, 2016.

\bibitem{jointcascade_fast}
H.~Qin, J.~Yan, X.~Li, and X.~Hu.
\newblock Joint training of cascaded cnn for face detection.
\newblock In {\em Computer Vision and Pattern Recognition (CVPR), 2016 IEEE
  Conference on}, 2016.

\bibitem{gabor_super}
S.~Ram~Dogiwal, Y.~Shishodia, and A.~Upadhyaya.
\newblock Super resolution image reconstruction using wavelet lifting schemes
  and gabor filters.
\newblock In {\em Confluence The Next Generation Information Technology Summit
  (Confluence), 2014 5th International Conference}. IEEE, 2014.

\bibitem{dpm_1}
R.~Ranjan and R.~Patel, Vishal M.and~Chellappa.
\newblock A deep pyramid deformable part model for face detection.
\newblock In {\em Biometrics Theory, Applications and Systems (BTAS), 2015 IEEE
  7th Int. Conf}, pages 1--8, 2015.

\bibitem{gabor_txt}
S.~Sabari~Raju, P.~Basa~Pati, and A.~Ramakrishnan.
\newblock Gabor filter based block energy analysis for text extraction from
  digital document images.
\newblock In {\em Document Image Analysis for Libraries, 2004. Proceedings.
  First International Workshop on}. IEEE, 2004.

\bibitem{vgg}
K.~Simonyan and A.~Zisserman.
\newblock Very deep convolutional networks for large-scale image recognition.
\newblock {\em CoRR}, abs/1409.1556, 2014.

\bibitem{fusion}
C.~Szegedy, V.~Vanhoucke, S.~Ioffe, J.~Shlens, and Z.~Wojna.
\newblock Rethinking the inception architecture for computer vision.
\newblock In {\em The IEEE Conference on Computer Vision and Pattern
  Recognition (CVPR)}, June 2016.

\bibitem{Tang}
Y.~Tang.
\newblock Deep learning using support vector machines.
\newblock {\em CoRR}, abs/1306.0239, 2013.

\bibitem{viola}
P.~Viola and M.~J.~Jones.
\newblock Robust real-time face detection.
\newblock {\em International journal of computer vision}, 57(2):137--154, 2004.

\bibitem{age_1}
T.~Wu and R.~Chellappa.
\newblock Age invariant face verification with relative craniofacial growth
  model.
\newblock In {\em European Conference on Computer Vision ECCV 2012: Computer
  Vision, ECCV 2012}, volume 7577, pages 58--71, 2012.

\bibitem{hybridmulti}
J.~Xing, K.~Li, W.~Hu, C.~Yuan, and H.~Ling.
\newblock Diagnosing deep learning models for high accuracy age estimation from
  a single image.
\newblock {\em Pattern Recognit}, 66:106--116, 2017.

\bibitem{faceness}
S.~Yang, P.~Luo, and X.~Change~Loy, Chenand~Tang.
\newblock From facial parts responses to face detection: A deep learning
  approach.
\newblock In {\em IEEE International Conference on Computer Vision}, pages
  3676--3684. IEEE, 2015.

\bibitem{widerface}
S.~Yang, P.~Luo, and X.~Change~Loy, Chenand~Tang.
\newblock Wider face: A face detection benchmark.
\newblock In {\em Computer Vision and Pattern Recognition (CVPR), 2016 IEEE
  Conference on}. IEEE, 2016.

\bibitem{casia}
D.~Yi, Z.~Lei, S.~Liao, and S.~Z. Li.
\newblock Learning face representation from scratch.
\newblock {\em CoRR}, abs/1411.7923, 2014.

\bibitem{yu}
Z.~Yu and C.~Zhang.
\newblock Image based static facial expression recognition with multiple deep
  network learning.
\newblock November 2015.

\bibitem{survey_facedetect}
S.~Zafeiriou, C.~Zhang, and Z.~Zhang.
\newblock A survey on face detection in the wild: Past, present and future.
\newblock {\em Computer Vision and Image Understandings}, 138:1--24, 2015.

\bibitem{MTCNN}
K.~Zhang, Z.~Zhang, Z.~Li, and Y.~Qiao.
\newblock Joint face detection and alignment using multi-task cascaded
  convolutional networks.
\newblock {\em IEEE Signal Processing Letters}, 23(10):1499--1503, 2016.

\end{thebibliography}

\end{document}